\documentclass[conference]{IEEEtran}
\IEEEoverridecommandlockouts
\usepackage{cite}
\usepackage{amsmath,amssymb,amsfonts}
\usepackage{algorithmic}
\usepackage{graphicx}
\usepackage{textcomp}
\usepackage{xcolor}
\usepackage{booktabs}
\def\BibTeX{{\rm B\kern-.05em{\sc i\kern-.025em b}\kern-.08em
    T\kern-.1667em\lower.7ex\hbox{E}\kern-.125emX}}
\begin{document}

\makeatletter
\newcommand{\linebreakand}{%
  \end{@IEEEauthorhalign}
  \hfill\mbox{}\par
  \mbox{}\hfill\begin{@IEEEauthorhalign}
}
\makeatother

\title{A Compositional Paradigm for Foundation Models: Towards Smarter Robotic Agents 
}

\author{%
\IEEEauthorblockN{%
Luigi Quarantiello\IEEEauthorrefmark{1}\IEEEauthorrefmark{4}, Elia Piccoli\IEEEauthorrefmark{1}, Jack Bell\IEEEauthorrefmark{1}, Malio Li\IEEEauthorrefmark{1}, Giacomo Carfì\IEEEauthorrefmark{1}, Eric Nuertey Coleman\IEEEauthorrefmark{1}}
\linebreakand
\IEEEauthorblockN{%
Gerlando Gramaglia\IEEEauthorrefmark{1}, Lanpei Li\IEEEauthorrefmark{1}\IEEEauthorrefmark{3}, Mauro Madeddu\IEEEauthorrefmark{1}, Irene Testa\IEEEauthorrefmark{1}, Vincenzo Lomonaco\IEEEauthorrefmark{2}} 
\linebreakand
\IEEEauthorblockA{\IEEEauthorrefmark{3}ISTI-CNR, Pisa, Italy}
\IEEEauthorblockA{\IEEEauthorrefmark{1}University of Pisa, Italy}
\IEEEauthorblockA{\IEEEauthorrefmark{2}LUISS University, Rome, Italy}
\IEEEauthorblockA{\IEEEauthorrefmark{4}Corresponding author: luigi.quarantiello@phd.unipi.it}}

\maketitle

\begin{abstract}
The birth of Foundation Models brought unprecedented results in a wide range of tasks, from language to vision, to robotic control.
These models are able to process huge quantities of data, and can extract and develop rich representations, which can be employed across different domains and modalities.
However, they still have issues in adapting to dynamic, real-world scenarios without retraining the entire model from scratch.
In this work, we propose the application of Continual Learning and Compositionality principles to foster the development of more flexible, efficient and smart AI solutions.
\end{abstract}

\section{Introduction}

Recent advancements in Artificial Intelligence have led to the emergence of Foundation Models (FMs).
Such architectures \cite{unisim}, leveraging billions of parameters, terabytes of training data and the most powerful (and energy-hungry) machines in the world, are able to encode broad knowledge and skills in their hidden spaces.
Ultimately, these features are employed to solve the most diverse tasks, encompassing different modalities and objectives.
The most impressive results, in particular from the user standpoint, comes from the large language models powering different chatbots, which excel at web search, code/text generation and brainstorming.
They show an incredible fluency (at least for widely spoken languages) and a great capability in elaborating knowledge and information from multiple sources.
Beyond language, FMs have demonstrated superior capabilities in other settings too, such as Computer Vision (\textit{e.g.} ViT for image classification), and multi-model learning (\textit{e.g.} CLIP for joint image-text understanding).
This led to networks with richer representations, which are more flexible and applicable to multiple scenarios, ultimately showing better performance.

However, recent releases
show that noticeable parameters and complexity increases lead only to marginal performance improvements.
In this work, we propose a alternative vision on AI that encompasses Continual Learning (CL) and Compositionality, to enhance FMs with an increased adaptability and flexibility to address more complex, dynamic scenarios.

\section{Large Models for Robotics}
Given their strong performance across different tasks, Transformer-based models have been applied to robotics too.
Early efforts, such as Robotics Transformer (RT), proved
the feasibility of high-capacity, data-driven control policies.
RT-1 \cite{rt1} demonstrated scalable generalization from large, diverse real-robot datasets, while RT-2 \cite{rt2} injected internet-scale vision-language priors directly into action policies, establishing the ``\textit{foundation model paradigm}'' for robotic control. In order to further scale these agents, the researchers focused on different aspects.
A first key contribution was the release of Open X-Embodiment dataset \cite{openx}, that unifies demonstrations across 22 robot types to enable cross-embodiment transfer.
On top of that, a great effort has been put to in the development of benchmarks, e.g. Meta-World \cite{metaworld}, and realistic yet highly efficient, GPU accelerated simulators, like ManiSkill3 \cite{maniskill}.
These tools enable efficient training and systematic evaluation of generalization and multi-task performance across different robotics tasks.
More recently, given the increased computational power and huge breakthroughs of language models, a parallel---or perhaps orthogonal---line of research has focused on the models architecture. Some pioneering work \cite{cliport, hiveformer} tried to leverage CLIP's embeddings to align natural language instruction in the robotics domain. As pre-trained models evolved, research aimed at modifying architectures to have specific components that handle different data modalities. While scaling architectures to billions of parameters helped in achieving better agents performance, a huge effort was spent to adapt general pre-trained models to specific robotics tasks. This effort lead to the development of Vision Language Action (VLA) models, in particular OpenVLA \cite{openvla}. By fine-tuning a 7B-Llama backbone, the agent obtains a \textit{16.5\%} increase in absolute task success rate compared to RT-2-X.


\section{The Need for Continual Learning and Compositionality}

In the previous sections, we examined Transformer-based models and their most recent applications, particularly in robotics.
As it often happens in Computer Science, many of these improvements has been driven by scaling, 
increasing model size, training data, or both, at the cost of growing computational and energy demands.
This alone has yielded impressive results: language models are able to handle complex requests and process longer inputs, offering a more human-like experience and demonstrating reduced biases and improved safety,
while robotic agents have gained more complex, multi-skill capabilities.
However, we are at a point of diminishing returns, where enlarging models, datasets or computational resources is not enough to obtain noticeable results \cite{luo2024has}.

Furthermore, despite being trained on massive datasets, FMs still struggle to generalize across tasks over time, especially when data distributions shift.
Their performance relies on statistical correlations rather than deeper reasoning, so scaling data \cite{scaling} or using Parameter-Efficient Fine-Tuning methods only yields limited improvements.
Applying CL principles and methods to FMs can represent a viable solution to give such models an higher degree of flexibility, allowing for updates on the pre-training distribution and efficient adaptation to domain-specific tasks \cite{bell2025future}.

\begin{table}[t]
    \centering
    \caption{\footnotesize Image Classification task. Our composition of adapters greatly outperforms competitors, while being more efficient}
    \begin{tabular}{ccc}
        \toprule
        \textbf{Method} & \textbf{Accuracy (\%)} & \textbf{Training Time (s)} \\
        \midrule
        SD-LoRA & 47.56 \tiny{$\pm$} 3.77 & 318.22 \\
        InfLoRA & 36.02 \tiny{$\pm$} 3.26 & 196.89 \\
        HAM (\textit{Ours}) & \textbf{55.17 \tiny{$\pm$} 1.04} & \textbf{170.61} \\
    \end{tabular}
    \vspace{-4mm}
    \label{tab:ham}
\end{table}
For these reasons, we argue that advancing to the next generation of AI requires a paradigm shift towards systems that are more flexible, adaptable and computationally efficient.
Compositionality---\textit{divide \& conquer}---is a key enabling factor of the next revolution.
It allows multiple specialized models to coexist, share knowledge and collectively solve complex tasks more effectively than single models.
Such model compositions are inherently dynamic; by changing the way in which the models are orchestrated, such systems could easily adapt to ever-changing scenarios, being more flexible than a monolithic solution.
Lastly, such combinations may be performed with little to no training phases, exploiting FMs full potential while being more efficient than the static counterparts.



\section{Experimental Results}
\begin{table}[t]
    \centering
    \caption{Robotic manipulation task. Our method outperforms more complex baselines, with reduced training times}
    \begin{tabular}{cccc}
        \toprule
        \textbf{Method} & \textbf{Reward/Step} & \textbf{Success Rate} & \textbf{Training Time} \\
        \midrule
        InstructRL & 0.14 & 0.0 & 40 h \\
        OpenVLA & 0.10 & 0.0 & 92 h \\
        WSA (\textit{Ours}) & \textbf{0.60} & \textbf{0.91} & \textbf{14 h} \\
    \end{tabular}
    \vspace{-6mm}
    \label{tab:wsa}
\end{table}

We explored the idea of CL and Compositionality in different domains, performing some preliminary experiments.

For image classification tasks, we studied the adaptation of the ViT model to a stream of tasks with the composition of multiple LoRA adapters \cite{hu2022lora}.
For each experience, we train a separate module, which is then merged with previously learned LoRAs to obtain a single final model.
We employ a hierarchical merging process: a first combination is performed within LoRAs grouped by similarity, while group adapters are later merged into a single module, so to have a unified model capable of solving every seen task \cite{coleman2025ham}.
We found that such method is effective in reducing the merging process interference.
Table \ref{tab:ham} shows the results obtained on CUB200 dataset with 50 tasks: our methods achieves an higher accuracy, while being faster in the training process.

In the context of robotics applications, we propose a simple architecture that using small adapters and attention mechanism can dynamically scale agents' pre-trained components while adapting to different contexts \cite{wsa}.
The results show that agents using small off-the-shelf pre-trained models and a small computational budget are able to learn to effectively solve robotics manipulation task while more complex counterparts as OpenVLA fail (given the same computational constraints).
The results obtained are presented in Table \ref{tab:wsa}, where its clear that our method, while being much more lightweight, obtains an higher reward per step on average, and achieves a far superior success rate than the competitors.


\section{Conclusion}
For FMs to reach their full potential and approach agentic AI, they must continuously adapt to evolving scenarios and generalize skills in open world contexts.
In this work, we show that composing multiple models enables strong performance with limited resources. 
Robotic agents must learn low level skills from both data and their interactions with the world in order to perform precise and specific behaviors. At the same time, agents need to be able to abstract from these policies and reason over their capabilities in order to come up with plans \cite{pi05}.
Learning a precise model of the world is key to understand how to interact with objects and humans, to continuously evolve agents abilities and adapt to sudden changes in the world without catastrophic forgetting and dangerous behaviors from safety misalignment.



\end{document}